\PassOptionsToPackage{table,xcdraw}{xcolor}
\documentclass[sigconf]{acmart}
\usepackage{graphicx}
\usepackage{microtype}
\usepackage{amsmath}
\usepackage{amsfonts}
\usepackage{multirow}
\usepackage{booktabs}
\usepackage{xspace}
\usepackage{color}
\usepackage{enumitem}
\usepackage{bbding}
\AtBeginDocument{%
  \providecommand\BibTeX{{%
    \normalfont B\kern-0.5em{\scshape i\kern-0.25em b}\kern-0.8em\TeX}}}

\copyrightyear{2022}
\acmYear{2022}
\setcopyright{acmcopyright}
\acmConference[MM '22] {Proceedings of the 30th ACM International Conference on Multimedia }{October 10--14, 2022}{Lisboa, Portugal.}
\acmBooktitle{Proceedings of the 30th ACM International Conference on Multimedia (MM '22), October 10--14, 2022, Lisboa, Portugal}
\acmPrice{15.00}
\acmISBN{978-1-4503-9203-7/22/10}
\acmDOI{10.1145/3503161.3547948}

\acmSubmissionID{814}

\begin{document}

\title[Enhancing Medical Vision-and-Language Pre-training with Knowledge]{Align, Reason and Learn: Enhancing Medical Vision-and-Language Pre-training with Knowledge}

\newlength{\mylen}
\settowidth{\mylen}{Shenzhen Research Institute of Big Data,}
\newlength{\mysecondlen}
\settowidth{\mysecondlen}{The Chinese University of Hong Kong,}

\author{Zhihong Chen}
\affiliation{
  \institution{\makebox[\mylen][c]{Shenzhen Research Institute of Big Data,}\\\makebox[\mylen][c]{The Chinese University of Hong Kong,}}
  \city{Shenzhen}
  \country{China}}
\email{zhihongchen@link.cuhk.edu.cn}

\author{Guanbin Li*}
\affiliation{
  \institution{Sun Yat-sen University,}
  \city{Guangzhou}
  \country{China}}
\email{liguanbin@mail.sysu.edu.cn}

\author{Xiang Wan*}
\affiliation{
  \institution{\makebox[\mylen][c]{Shenzhen Research Institute of Big Data,}\\\makebox[\mylen][c]{The Chinese University of Hong Kong,}}
  \city{Shenzhen}
  \country{China}\\
  \institution{Pazhou Lab,}
  \city{Guangzhou}
  \country{China}}
\email{wanxiang@sribd.cn}

\makeatletter
\def\authornotetext#1{
	\if@ACM@anonymous\else
	\g@addto@macro\@authornotes{
		\stepcounter{footnote}\footnotetext{#1}}
	\fi}
\makeatother
\authornotetext{Corresponding authors.}






\def\authors{Zhihong Chen, Guanbin Li, Xiang Wan}

\renewcommand{\shortauthors}{Zhihong Chen, Guanbin Li, \& Xiang Wan}

\begin{abstract}
Medical vision-and-language pre-training (Med-VLP) has received considerable attention owing to its applicability to extracting generic vision-and-language representations from medical images and texts.
Most existing methods mainly contain three elements: uni-modal encoders (i.e., a vision encoder and a language encoder), a multi-modal fusion module, and pretext tasks, with few studies considering the importance of medical domain expert knowledge and explicitly exploiting such knowledge to facilitate Med-VLP.
Although there exist knowledge-enhanced vision-and-language pre-training (VLP) methods in the general domain, most require off-the-shelf toolkits (e.g., object detectors and scene graph parsers), which are unavailable in the medical domain.
In this paper, we propose a systematic and effective approach to enhance Med-VLP by structured medical knowledge from three perspectives.
First, considering knowledge can be regarded as the intermediate medium between vision and language, we align the representations of the vision encoder and the language encoder through knowledge.
Second, we inject knowledge into the multi-modal fusion model to enable the model to perform reasoning using knowledge as the supplementation of the input image and text.
Third, we guide the model to put emphasis on the most critical information in images and texts by designing knowledge-induced pretext tasks.
To perform a comprehensive evaluation and facilitate further research, we construct a medical vision-and-language benchmark including three tasks.
Experimental results illustrate the effectiveness of our approach, where state-of-the-art performance is achieved on all downstream tasks.
Further analyses explore the effects of different components of our approach and various settings of pre-training.\footnote{The source code is available at~\url{https://github.com/zhjohnchan/ARL}.}
\end{abstract}



\maketitle

\section{Introduction}
Medical data streams from various sources, among which vision and language are two critical ones.
It includes image data (e.g., radiography, magnetic resonance imaging, and computed tomography) and text data (e.g., radiology reports and medical texts).
Medical vision-and-language pre-training (Med-VLP) aims to jointly process data from these two modalities to learn generalizable multi-modal representations from large-scale medical image-text data.
It enables a vision-and-language model to address a wide range of medical vision-and-language tasks (e.g., medical visual question answering (Med-VQA), medical image-text classification (Med-ITC), and medical image-text retrieval (Med-ITR)), which can be crucial for alleviating the data scarcity problem in the medical field.

In the past few years, vision-and-language pre-training (VLP) has drawn sustaining attention \cite{chen2020uniter,kim2021vilt,su2019vlbert,tan2019lxmert,lu2019vilbert,li2020oscar,cui2021rosita} and achieved state-of-the-art performance on many vision-and-language tasks in the general domain.
In general, a VLP system consists of three elements: (i) uni-modal encoders (i.e., a vision encoder and a language encoder) that encode images and texts into image and text features, respectively; (ii) a multi-modal fusion module that performs the fusion of the encoded image and text features; (iii) pretext tasks (e.g., masked image modeling (MIM), masked language modeling (MLM), and image-text matching (ITM)) that assist the learning of VLP models.
More recently, some studies \cite{li2020comparison,khare2021mmbert,gong2021mtpt,moon2021multi}  applied VLP to the medical domain and significantly improved the performance for medical vision-and-language tasks (especially for Med-VQA).
These methods are superior in capturing the mappings between images and texts and thus enable the pre-trained models to understand the complicated cross-modal information.
For example, \cite{khare2021mmbert,gong2021mtpt} proposed to perform the pre-training on medical image-text pairs to capture medical knowledge, and the evaluation on Med-VQA has demonstrated the validity of their proposed methods.

Although these methods have motivated the learning of image-text correspondences through well-designed model architectures and pretext tasks, most of them disregard the complementary information (i.e., knowledge) shared by different modalities and still lack the explicit knowledge modeling for Med-VLP.
Even in the general domain, there are only a few VLP studies \cite{li2020oscar,yu2021ernie-vil,chen2021kb-vlp,cui2021rosita} on incorporating external knowledge into the pre-training process.
For instance, ERNIE-ViL \cite{yu2021ernie-vil} constructed a scene graph from the input text to build the semantic connections between vision and language and emphasized the importance of keywords (e.g., objects, attributes, and relationships between objects) through the designs of pretext tasks.
ROSITA \cite{cui2021rosita} used a unified scene graph shared by the input image and text to enhance the
semantic alignments between vision and language.
Similarly, KB-VLP \cite{chen2021kb-vlp} used object tags detected from images and knowledge graph embeddings extracted from texts to enhance the learning of knowledge-aware representations.
However, the aforementioned studies require off-the-shelf toolkits (e.g., object detectors and scene graph parsers), which are generally unavailable in the medical domain.
Furthermore, they might be limited in scalability as their performance depends heavily on the reliability of the object detectors or scene graph parsers.
Therefore, it is expected to have a better solution to exploit external knowledge more appropriately and systematically and further improve the generalization ability of Med-VLP methods.

In this paper, we propose a systematic approach to Med-VLP enhanced by structured expert domain knowledge from the Unified Medical Language System \cite{bodenreider2004umls} (UMLS), a large medical knowledge base containing many biomedical terminologies with the associated information, such as synonyms and categorical groupings.
To ensure the effectiveness and efficiency of our approach, structured knowledge is injected into the Med-VLP system from three perspectives:
(i) Aligning Through Knowledge: It uses knowledge as the intermediate medium between vision and language to align the image and text features encoded by the uni-modal encoders;
(ii) Reasoning Using Knowledge: It develops a knowledge-enhanced multi-modal fusion module to integrate knowledge into the interaction process of the image and text features;
(iii) Learning From Knowledge: It constructs knowledge-induced pretext tasks to assist the model in capturing underlying critical medical information of the images and texts to promote the medical vision-and-language understanding.
As a result, the proposed method is able to learn cross-modal domain-specific knowledge from large-scale medical image-text datasets and medical knowledge bases to promote the learning of semantically aligned and knowledge-aware image and text representations.
We perform the pre-training on three large-scale medical image-text datasets, i.e., ROCO \cite{pelka2018roco}, MedICaT \cite{subramanian2020medicat}, and MIMIC-CXR \cite{johnson2019mimic}.
To verify the effectiveness of our approach and facilitate further research, we construct a medical vision-and-language understanding benchmark including three tasks (i.e., Med-VQA, Med-ITC, and Med-ITR).
Experimental results demonstrate the effectiveness of our approach, where state-of-the-art performance is achieved on all datasets.
%
\begin{figure*}[t]
\centering
\includegraphics[width=0.95\textwidth, trim=0 0 0 0]{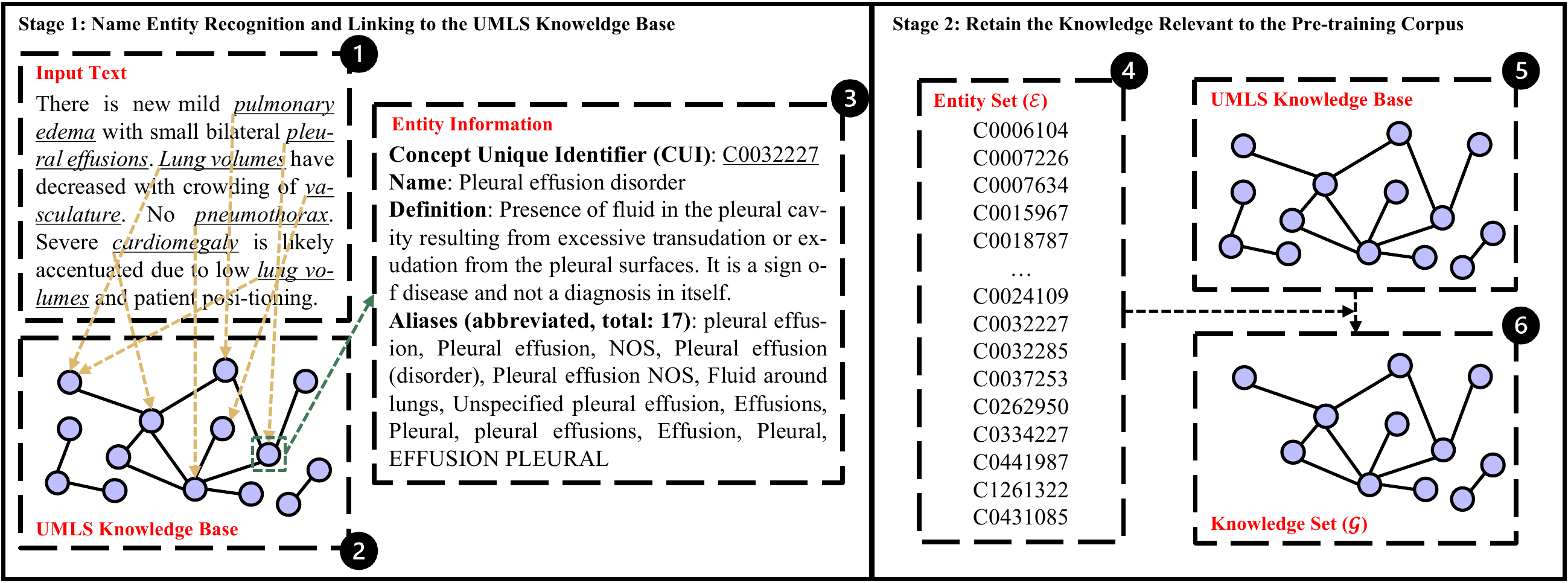}
\caption{The flowchart of knowledge extraction from UMLS, a medical knowledge base.
It contains two main stages, where the first stage is to link entities of each input text to the knowledge base and the second stage is to retain the knowledge relevant to the pre-training corpus to form our knowledge set.
Numbers are marked for ease of reading.}
\label{fig:knowledge-extraction}
\end{figure*}

\section{Related Work}
\noindent\textbf{Vision-and-Language Pre-training (VLP)}~
Motivated by the success of the self-supervised pre-training recipe of BERT in NLP, there has been an increasing interest in developing VLP methods to address a wide range of vision-and-language tasks.
In general, VLP methods can be categorized with respect to three perspectives.
For the designs of the uni-modal encoders, different methods adopt different image features (e.g., region features \cite{li2019visualbert,lu2019vilbert}, patch embeddings \cite{kim2021vilt,li2021albef,wang2021simvlm}, and grid features \cite{huang2020pixel}) and distinct text features (e.g., statistic embeddings \cite{kim2021vilt} and dynamic embeddings \cite{dou2021meter}).
For multi-modal fusion modules, existing methods can be classified into two categories (i.e., single-stream and dual-stream).
In specific, for the single-stream fusion, the models \cite{li2019visualbert,chen2020uniter,su2019vlbert,li2020oscar} use a single Transformer for early and unconstrained fusion between modalities; for the dual-stream fusion, the models \cite{tan2019lxmert,lu2019vilbert,yu2021ernie-vil} adopt the co-attention mechanism to interact different modalities.
For pretext tasks, inspired by uni-modal pre-training schemes such as MLM \cite{devlin2019bert,liu2019roberta} and causal language modeling \cite{brown2020gpt-3}, existing studies explore a variety of pre-training tasks, including MLM \cite{li2019visualbert,lu2019vilbert,tan2019lxmert}, MIM \cite{lu2019vilbert,chen2020uniter}, ITM \cite{li2019visualbert,zhang2021vinvl}, image-text contrastive \cite{li2021albef} and prefix language modeling \cite{wang2021simvlm}.
This paper adopts a purely Transformer-based backbone architecture using the dual-stream fusion with ViT-based grid features and BERT-based dynamic text features and three common pretext tasks (i.e., MLM, MIM, and ITM).

\noindent\textbf{Medical Vision-and-Language Pre-training (Med-VLP)}~
Being one of the applications and extensions of VLP to the medical domain, Med-VLP aims to understand the content and relations between medical images and their corresponding texts.
It can be traced back to \cite{li2020comparison}, which explored the performance of four vision-and-language models pre-trained in the general domain on a disease classification task.
Then MMBERT \cite{khare2021mmbert}, PubMedCLIP \cite{eslami2021does}, and MedViLL \cite{moon2021multi} performed pre-training on medical image-text data before fine-tuning on the downstream tasks.
Compared with these studies, we design a more appropriate and systematic scheme for Med-VLP from four aspects (i.e., pre-training datasets, model designs, pre-training tasks, and evaluation benchmarks).

\noindent\textbf{Knowledge-Enhanced Pre-training}~
For uni-modal pre-training in CV and NLP, many works have investigated how to incorporate knowledge into the pre-trained models.
According to the knowledge injection schemes, existing studies can be classified into four categories: embeddings combination \cite{zhang2019ernie,peters2019knowbert}, data structure compatibility \cite{liu2020k-bert,sun2020colake,he2020bert-mk}, knowledge supervision \cite{sun2019ernie-baidu,wang2021kepler}, and neural-symbolic methods \cite{amizadeh2020neuro}.
For VLP, knowledge can be acquired from both the image and text modalities, and there are several works \cite{li2020oscar,yu2021ernie-vil,cui2021rosita} studying to integrate knowledge into their methods.
ERNIE-ViL \cite{yu2021ernie-vil} built detailed semantic alignments between vision and language based on the scene graph parsed from the text.
ROSITA \cite{cui2021rosita} proposed integrating extra cross-modal knowledge mappings to enhance the learning of semantic alignments between vision and language.
Different from them, we revisit existing knowledge-enhanced methods and propose to inject knowledge from three VLP-specific perspectives without requiring object detectors or scene graph parsers, which are unavailable in the medical domain.

\section{The Proposed Approach}
\begin{figure*}[t]
\centering
\includegraphics[width=0.95\textwidth, trim=0 0 0 0]{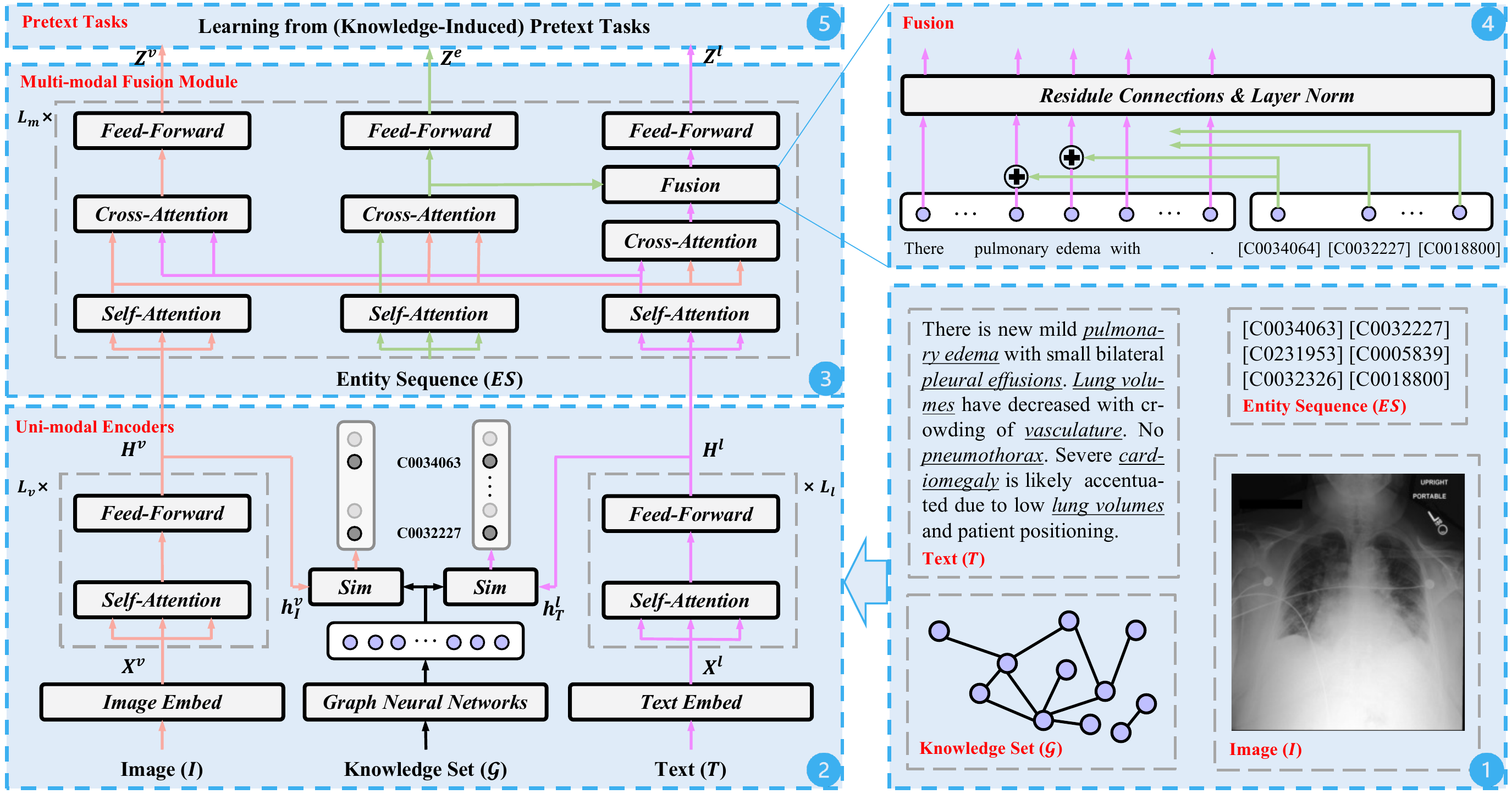}
\caption{The overall architecture of our proposed approach, where the inputs, uni-modal encoders (with the ``aligning through knowledge'' process), multi-modal fusion module (with the ``reasoning using knowledge'' process), pretext tasks (with the ``learning from knowledge'' process) are shown in blue dash boxes.
Numbers are marked for ease of reading.}
\label{fig:framework}
\end{figure*}

We follow the standard pre-train-and-fine-tune paradigm for medical vision-and-language understanding.
In the pre-training stage, the framework develops a variety of pretext tasks to train the Med-VLP model using medical image-text pairs.
In the fine-tuning stage, the pre-trained Med-VLP model is transferred to various medical vision-and-language downstream tasks.
An overview of the proposed approach is demonstrated in Figure \ref{fig:framework}, and the details of the general Med-VLP framework, the knowledge extraction process (shown in Figure \ref{fig:knowledge-extraction}), and the injection of knowledge into the general Med-VLP framework are introduced in the following subsections.

\subsection{The General Med-VLP Framework}
The general Med-VLP framework can be partitioned into three major components, i.e., the uni-modal encoders, the multi-modal fusion module, and the pretext tasks.
The overall description of the three components is detailed below.

\noindent\textbf{Uni-modal Encoders}~
In the Med-VLP framework, there is a vision encoder and a language encoder, which encode the input image and text into image and text features, respectively.

For the vision encoder, we study the use of vision Transformer \cite{dosovitskiy2020vit} (ViT).
In ViT, an input image $\boldsymbol{I} \in \mathbb{R}^{H \times W \times C}$ is first segmented into patches $\{\boldsymbol{x_{1}^{v}}, \boldsymbol{x_{2}^{v}}, \ldots, \boldsymbol{x_{N_{v}}^{v}}\}$, where $H \times W$ is the image resolution, $C$ is the number of channels, $N_{v}$ is the number of patches, $\boldsymbol{x_{i}^{v}} \in \mathbb{R}^{P^{2} \times C}$ and $P \times P$ is the patch resolution.
Then the patches $\{\boldsymbol{x_{1}^{v}}, \boldsymbol{x_{2}^{v}}, \ldots, \boldsymbol{x_{N_{v}}^{v}}\}$ are flattened and linearly projected into patch embeddings through a linear transformation $\boldsymbol{E^{v}} \in \mathbb{R}^{P^{2}C \times D}$ and a special learnable token embedding $\boldsymbol{x_{I}^{v}} \in \mathbb{R}^{D}$ is prepended for the aggregation of visual information.
Therefore, the input image representations are obtained via summing up the patch embeddings and learnable 1D position embeddings $\boldsymbol{E^{v}_{pos}} \in \mathbb{R}^{(N_{v}+1) \times D}$:
\begin{equation}
    \boldsymbol{X^{v}} = [\boldsymbol{x_{I}^{v}}; \boldsymbol{x_{1}^{v}}\boldsymbol{E^{v}}; \boldsymbol{x_{2}^{v}}\boldsymbol{E^{v}}; \ldots; \boldsymbol{x_{N_{v}}^{v}}\boldsymbol{E^{v}}] + \boldsymbol{E^{v}_{pos}}.
\end{equation}
Then $\boldsymbol{X^{v}}$ is fed into a Transformer model with $L_{v}$ Transformer layers.
Finally, we obtain the contextualized image representations $ \boldsymbol{H^{v}} = [\boldsymbol{h^{v}_{I}}; \boldsymbol{h^{v}_{1}}; \boldsymbol{h^{v}_{2}}; ...; \boldsymbol{h^{v}_{N_{v}}}]$.

For the language encoder, we follow BERT \cite{devlin2019bert} to tokenize the input text to subword tokens $\{x_{1}^{l},x_{2}^{l},\ldots,x_{N_{l}}^{l}\}$ by WordPiece \cite{wu2016google} and then represent subword tokens as $\{\boldsymbol{x_{1}^{l}},\boldsymbol{x_{2}^{l}},\ldots,\boldsymbol{x_{N_{l}}^{l}}\}$, where $\boldsymbol{x_{i}^{l}} \in \mathbb{R}^{V}$ are the one-hot form of $x_{i}^{l}$, $V$ is the vocabulary size, and $N_{l}$ is the number of tokens.
Subsequently, the tokens are linearly projected into embeddings through a linear transformation $\boldsymbol{E^{l}} \in \mathbb{R}^{V\times D}$. 
Afterwards, a start-of-sequence token embedding $\boldsymbol{x_{T}^{l}} \in \mathbb{R}^{D}$ and a special boundary token embedding $\boldsymbol{x_{SEP}^{l}} \in \mathbb{R}^{D}$ are added to the text sequence.
Therefore, the input text representations are computed via summing up the token embeddings and text position embeddings $\boldsymbol{E^{l}_{pos}} \in \mathbb{R}^{(N_{l}+2) \times D}$:
\begin{equation}
    \boldsymbol{X^{l}} = [\boldsymbol{x_{T}^{l}}; \boldsymbol{x_{1}^{l}E^{l}}; \ldots; \boldsymbol{x_{N_{l}}^{l}E^{l}}; \boldsymbol{{x}_{SEP}^{l}}] + \boldsymbol{E^{l}_{pos}}.
\end{equation}
Then $\boldsymbol{X^{l}}$ is fed into a Transformer model with $L_{l}$ Transformer layers.
Finally, we obtain the contextualized text representations $\boldsymbol{H^{l}} = [\boldsymbol{h^{l}_{T}}; \boldsymbol{h^{l}_{1}}; \boldsymbol{h^{l}_{2}}; \ldots; \boldsymbol{h^{l}_{N_{l}}}; \boldsymbol{h^{l}_{SEP}}]$.

\noindent\textbf{Multi-model Fusion Module}~
We adopt the co-attention mechanism in the multi-modal fusion module to fuse the contextualized representations from images and texts.
In detail, the multi-modal fusion module consists of two Transformer models for vision and language, respectively, each of which is a stack of $L_{m}$ Transformer layers.
In each Transformer layer, there are three sub-layers, i.e., a self-attention sub-layer, a cross-attention sub-layer, and a feed-forward sub-layer.
The attention mechanism is applied in the self-attention and cross-attention sub-layers and is defined as
\begin{equation}
    \text{ATTN}(\boldsymbol{Q}, \boldsymbol{K}, \boldsymbol{V})=\operatorname{softmax}\left(\boldsymbol{Q} \boldsymbol{K}^{\top}/\sqrt{D_{k}}\right) \cdot \boldsymbol{V},
\end{equation}
where $\boldsymbol{Q}$, $\boldsymbol{K}$, and $\boldsymbol{V}$ are the query, key, value matrices linearly transformed from the corresponding input sequences, respectively, and $D_{k}$ is the dimension of $\boldsymbol{K}$.
In the self-attention sub-layer, the representations interact within modalities:
\begin{align}
    \boldsymbol{H^{vs}}&=\text{ATTN}(\boldsymbol{H^{v}}, \boldsymbol{H^{v}}, \boldsymbol{H^{v}}),\\
    \boldsymbol{H^{ls}}&=\text{ATTN}(\boldsymbol{H^{l}}, \boldsymbol{H^{l}}, \boldsymbol{H^{l}}),
\end{align}
where $\boldsymbol{H^{vs}}$ and $\boldsymbol{H^{ls}}$ are the self-attention outputs for vision and language, respectively.
Then residual connections followed by layer normalization are employed to $\boldsymbol{H^{vs}}$ and $\boldsymbol{H^{ls}}$ and we denote the results as $\boldsymbol{H^{vs}}$ and $\boldsymbol{H^{ls}}$, respectively, for simplicity.
In the cross-attention sub-layer, the representations interact across modalities to integrate cross-modal information into their representations:
\begin{align}
    \boldsymbol{H^{vc}}=\text{ATTN}(\boldsymbol{H^{vs}}, \boldsymbol{H^{ls}}, \boldsymbol{H^{ls}}),\\
    \boldsymbol{H^{lc}}=\text{ATTN}(\boldsymbol{H^{ls}}, \boldsymbol{H^{vs}}, \boldsymbol{H^{vs}}),
    \label{equation:cross-attention}
\end{align}
where $\boldsymbol{H^{vc}}$ and $\boldsymbol{H^{lc}}$ are the cross-attention outputs for vision and language, respectively.
Similarly, residual connections followed by layer normalization are employed to $\boldsymbol{H^{vc}}$ and $\boldsymbol{H^{lc}}$, and we denote the results as $\boldsymbol{H^{vc}}$ and $\boldsymbol{H^{lc}}$, respectively, for simplicity.
Finally, $\boldsymbol{H^{vc}}$ and $\boldsymbol{H^{lc}}$ are input to the feed-forward sub-layer (i.e., a multi-layer perceptron (MLP)) to obtain the multi-modal representations $\boldsymbol{Z^{v}} = [\boldsymbol{z^{v}_{I}}; \boldsymbol{z^{v}_{1}}; \boldsymbol{z^{v}_{2}}; \ldots; \boldsymbol{z^{v}_{N_{v}}}]$ for vision and $\boldsymbol{Z^{l}} = [\boldsymbol{z^{l}_{T}}; \boldsymbol{z^{l}_{1}}; \boldsymbol{z^{l}_{2}}; \ldots; \boldsymbol{z^{l}_{N_{l}}}; \boldsymbol{z^{l}_{SEP}}]$ for language.

\noindent\textbf{Pretext Tasks}~
Given the aforementioned structure (denoted as $\mathcal{M}_{\theta}$) with its parameters $\theta$, the Med-VLP framework develops various pretext tasks (e.g., masked language modeling (MLM), masked image modeling (MIM), and image-text matching (ITM)) to guide the learning of $\theta$.
Assuming there are $S$ pretext tasks, the learning process can be formalized as
\begin{equation}
    \theta^{*},\theta_{1}^{*},...,\theta_{S}^{*} = \mathop{\arg\min}_{\theta,\theta_{1},...,\theta_{S}}
    \sum_{i=1}^{S} L_{i}
    (Y_{i}, \mathcal{D}_{\theta_{i}}(\mathcal{M}_{\theta}(I, T)),
\end{equation}
where $L_{i}$ are the loss functions of pretext tasks, $Y_{i}$ are the corresponding ground-truth labels, and $\mathcal{D}_{\theta_{i}}$ are the prediction heads with their parameters $\theta_{i}$.

\subsection{Knowledge Extraction}
Although knowledge graphs (KGs) have shown their effectiveness in many natural language processing (NLP) tasks \cite{zhang2019ernie,peters2019knowbert,liu2020k-bert} and computer vision (CV) tasks \cite{marino2017more,wang2018zero}, the existing Med-VLP methods rarely consider incorporating KGs to provide rich structured knowledge for better vision-and-language understanding.

Therefore, we propose to enhance Med-VLP by leveraging external domain expert knowledge from UMLS.
In doing so, we extract knowledge through two stages, as illustrated in Figure \ref{fig:knowledge-extraction}.
The first stage is to apply a named entity recognition, and linking tool ScispaCy \cite{neumann2019scispacy} to pre-process the texts in the pre-training corpus to link entities in the texts to the UMLS knowledge base for entity disambiguation.
Therefore, for each image-text pair, there is an entity sequence $ES=\{x_{1}^{e},x_{2}^{e},\ldots,x_{N_{es}}^{e}\}$ aligning to the token sequence $T=\{x_{1}^{l},x_{2}^{l},\ldots,x_{N_{l}}^{l}\}$, where $x_{i}^{e}$ are the extracted entities and $N_{es}$ is the length of the entity sequence.
Following \cite{diao2020zen}, to record the position of the extracted entities, we adopt an entity matching matrix $\boldsymbol{P} \in \mathbb{R}^{N_{l} \times N_{es}}$, where each element is represented by
\begin{equation}
    \boldsymbol{P}_{i j}= \begin{cases}1 & x_{i}^{l} \in x_{j}^{e} \\ 0 & x_{i}^{l} \notin x_{j}^{e}\end{cases},
\end{equation}
where $P$ is employed to assist the interaction between the text and the entity sequence in the knowledge injection process (as described in the next subsection).
After pre-processing all the texts, we can obtain an entity set $\mathcal{E}=\{e_{i}\}_{i=1}^{N_{e}}$ containing all the $N_{e}$ entities related to the pre-training corpus.
The second stage is to extract relevant knowledge graph triples from the UMLS knowledge base once both the head and tail entities of the triple are in the entity set $\mathcal{E}$.
We denote the extracted knowledge graph (i.e., a sub-graph of the UMLS knowledge base) as the knowledge set $\mathcal{G}=\{k_{i}=(h_{i}, r_{i}, t_{i})\}_{i=1}^{N_{g}}$, where $N_{g}$ is the number of knowledge graph triples, $k_{i}$ are the knowledge graph triples, and $h_{i}$, $r_{i}$ and $t_{i}$ represent the head entity, relation, and tail entity, respectively.

\subsection{Knowledge Injection}
To integrate knowledge into the general Med-VLP framework, first, we perform knowledge representation following two steps:
(i) We apply knowledge representation learning algorithm (e.g., TransE \cite{bordes2013transe}) to the knowledge graph $\mathcal{G}=\{k_{i}=(h_{i}, r_{i}, t_{i})\}_{i=1}^{N_{g}}$ to obtain the entity embeddings $\{\boldsymbol{e_{i}}\}_{i=1}^{N_{e}}$, where $\boldsymbol{e_{i}} \in \mathbb{R}^{D_{e}}$ and $D_{e}$ is the dimension of the entity embeddings;
(ii) We adopt Graph Neural Networks (e.g., Graph Attention Networks \cite{velivckovic2018gat}) to take account of the whole structure of the graph to aggregate local information in the graph neighborhood for each node, and obtain the entity representations (denoted as $\{\boldsymbol{e_{i}}\}_{i=1}^{N_{e}}$ for simplicity, where $\boldsymbol{e_{i}} \in \mathbb{R}^{D_{e}}$).

Afterwards, given the input image $I$ and text $T=\{x_{1}^{l},x_{2}^{l},\ldots,x_{N_{l}}^{l}\}$ with its corresponding entity sequence $ES=\{x_{1}^{e},x_{2}^{e},\ldots,x_{N_{es}}^{e}\}$, we develop three essential and systematic designs to inject knowledge from the following perspectives:

\noindent\textbf{(i) Aligning Through Knowledge}~
Knowledge can be regarded as the intermediate medium between vision and language, where knowledge can be used as an explanation of the meaning behind both images and texts.
In most cases, entities in knowledge graph triples can be treated as anchor points that appear in the image and are mentioned in the accompanying text.
Motivated by this fact, we propose to align the image representations and the text representations from uni-modal encoders through knowledge.
Similar to \cite{li2021albef}, it serves two purposes:
It improves the unimodal encoders to better understand the semantic meaning of images and texts;
It eases the learning of semantic alignments between images and texts.

Formally, given the aggregated image representation $\boldsymbol{h^{v}_{I}}$ and text representation $\boldsymbol{h^{l}_{T}}$, we calculate the image-knowledge and text-knowledge similarity followed by a sigmoid function:
\begin{align}
    p_{i}^{v} &= \text{sigmoid}(\boldsymbol{e_{i}}^{\top}\boldsymbol{W_{vk}}\boldsymbol{h^{v}_{I}}), i=1,...,N_{e},\\
    p_{i}^{l} &= \text{sigmoid}(\boldsymbol{e_{i}}^{\top}\boldsymbol{W_{lk}}\boldsymbol{h^{l}_{T}}), i=1,...,N_{e},
\end{align}
where $\boldsymbol{W_{vk}} \in \mathbb{R}^{D_{e} \times D}$ and $\boldsymbol{W_{lk}} \in \mathbb{R}^{D_{e} \times D}$ are trainable weights for the linear transformation.
Therefore, the alignments for image-knowledge and text-knowledge are learned explicitly through minimizing the following functions:
\begin{align}
    L_{vk} &= -\sum_{i=1}^{N_{e}} ( y_{i} \log p_{i}^{v} + \left(1-y_{i}\right) \log \left(1-p_{i}^{v}\right) ), \\
    L_{lk} &= -\sum_{i=1}^{N_{e}} ( y_{i} \log p_{i}^{l} + \left(1-y_{i}\right) \log \left(1-p_{i}^{l}\right) ),
\end{align}
where $y_{i}$ can be defined as
\begin{equation}
    y_{i}= \begin{cases}1 & e_{i} \in ES \\ 0 & e_{i} \notin ES\end{cases}.
\end{equation}
Therefore, knowledge is employed as an intermediate medium to enhance and smooth image-text mappings by doing so.

\noindent\textbf{(ii) Reasoning Using Knowledge}~
As the supplementation of the input image and text, knowledge can also be utilized to assist the reasoning of the Med-VLP model.
In doing so, we enhance the multi-modal fusion module by knowledge.

Formally, given the entity sequence $ES=\{x_{1}^{e},x_{2}^{e},...,x_{N_{es}}^{e}\}$, first, we extract its entity representations $\boldsymbol{H^{e}} = [\boldsymbol{h^{e}_{1}}; \boldsymbol{h^{e}_{2}}; \ldots; \boldsymbol{h^{e}_{N_{es}}}]$.
Second, we apply self-attention to the entity representations to encode the contextualized information:
\begin{equation}
    \boldsymbol{H^{es}}=\text{ATTN}(\boldsymbol{H^{e}}, \boldsymbol{H^{e}}, \boldsymbol{H^{e}}),
\end{equation}
where $\boldsymbol{H^{es}}$ is the self-attention outputs of entity representations.
Residual connections followed by layer normalization are employed to $\boldsymbol{H^{es}}$, and we denote the outputs as $\boldsymbol{H^{es}}$ for simplicity.
Third, to interact the entities with the image, since there is no available toolkits to structuralize the input image to construct mappings between image patches and entities, we directly perform cross-attention on $\boldsymbol{H^{es}}$ and $\boldsymbol{H^{vs}}$:
\begin{equation}
    \boldsymbol{H^{ec}}=\text{ATTN}(\boldsymbol{H^{es}}, \boldsymbol{H^{vs}}, \boldsymbol{H^{vs}}),
\end{equation}
where $\boldsymbol{H^{ec}}$ is the cross-attention outputs of entity representations.
Residual connections followed by layer normalization are employed to $\boldsymbol{H^{ec}}$, and we denote the outputs as $\boldsymbol{H^{ec}}$ for simplicity.
Fourth, since the mappings between the entities and the text are recorded by $\boldsymbol{P}$, we can use it to fuse $\boldsymbol{H^{ec}}$ and $\boldsymbol{H^{ls}}$ through:
\begin{equation}
    \boldsymbol{\Tilde{H}^{lc}} = \boldsymbol{P}\boldsymbol{H^{ec}} + \boldsymbol{H^{lc}},
\end{equation}
where $\boldsymbol{\Tilde{H}^{lc}}$ is the text representations encoded with the image and entity information.
Finally, we apply residual connections followed by layer normalization to $\boldsymbol{\Tilde{H^{lc}}}$ and input it to the feed-forward sub-layer to complete the knowledge-enhanced multi-modal fusion.
In the meantime, $\boldsymbol{H^{ec}}$ is input to another feed-forward sub-layer to produce the representations $\boldsymbol{Z^{e}}$ for the next layer.

\noindent\textbf{(iii) Learning From Knowledge}~
Knowledge can help us to induce more sophisticated pretext tasks to guide the model to learn more informative representations.
In our paper, we follow \cite{sun2019ernie-baidu} to design a knowledge-induced mask generation strategy.
Specifically, when performing the MLM task given the input text $T=\{x_{1}^{l},x_{2}^{l},\ldots,x_{N_{l}}^{l}\}$ with its entity sequence $ES=\{x_{1}^{e},x_{2}^{e},\ldots,x_{N_{es}}^{e}\}$, we do not mask subword tokens in $T$ randomly.
Instead, we randomly sample entities from $ES$ and then mask consecutive spans of subword tokens belonging to the sampled entities.
Since entities can be abstract or have a physical existence, it can force the model to focus on critical medical information in both images and texts.

Therefore, knowledge can be injected into the Med-VLP framework in a systematic way through the above three designs.

\section{Experimental Settings}
\subsection{Pre-training Setup}
\begin{table*}[t]
\centering
\caption{Results on the Med-VQA task (including three datasets, i.e., VQA-RAD, SLACK, and VQA-2019) to compare with the state-of-the-art methods.
Dark and light grey colors highlight the top and second best results on each evaluation metric.}
\label{table:med-vqa}
\begin{tabular}{@{}ll|ccccccccc|c@{}}
\toprule
\multicolumn{2}{c|}{}                          & MFB   & SAN   & BAN   & MEVF-SAN & MEVF-BAN & CPRD-BAN                      & COND-REA                      & MTPT  & MMBERT                        &                               \\
\multicolumn{2}{c|}{\multirow{-2}{*}{Dataset}} & \cite{yu2017mfb} & \cite{yang2016san} & \cite{kim2018ban} & \cite{nguyen2019mevf} & \cite{nguyen2019mevf} & \cite{liu2021cprd} & \cite{zhan2020conditional} & \cite{gong2021mtpt} & \cite{khare2021mmbert}                           & \multirow{-2}{*}{Ours}        \\ \midrule
                                & Open         & 14.50 & 31.30 & 37.40 & 49.20    & 49.20    & 52.50                         & 60.00                         & 61.50 & \cellcolor[HTML]{EFEFEF}63.10 & \cellcolor[HTML]{C0C0C0}67.60 \\
                                & Closed       & 74.30 & 69.50 & 72.10 & 73.90    & 77.20    & 77.90                         & 79.30 & \cellcolor[HTML]{EFEFEF}80.90 & 77.90                         & \cellcolor[HTML]{C0C0C0}86.76 \\
\multirow{-3}{*}{VQA-RAD}       & Overall      & 50.60 & 54.30 & 58.30 & 64.10    & 66.10    & 67.80                         & 71.60                         & \cellcolor[HTML]{EFEFEF}73.20 & 72.00 & \cellcolor[HTML]{C0C0C0}79.16 \\ \midrule
                                & Open         & 72.20 & 74.00 & 74.60 & 75.30    & 77.80    & \cellcolor[HTML]{EFEFEF}79.50 & -                             & -     & -                             & \cellcolor[HTML]{C0C0C0}81.89 \\
                                & Closed       & 75.00 & 79.10 & 79.10 & 78.40    & 79.80    & \cellcolor[HTML]{EFEFEF}83.40 & -                             & -     & -                             & \cellcolor[HTML]{C0C0C0}91.35 \\
\multirow{-3}{*}{SLACK}         & Overall      & 73.30 & 76.00 & 76.30 & 76.50    & 78.60    & \cellcolor[HTML]{EFEFEF}81.10 & -                             & -     & -                             & \cellcolor[HTML]{C0C0C0}85.59 \\ \midrule
VQA-2019                        & Overall      & -     & -     & -     & 68.90    & 77.86    & -                             & -                             & -     & \cellcolor[HTML]{EFEFEF}77.90 & \cellcolor[HTML]{C0C0C0}80.32 \\ \bottomrule
\end{tabular}
\end{table*}
\begin{table}[t]
\centering
\caption{Results on the Med-ITC task (i.e., the MELINDA dataset) to compare with the state-of-the-art methods.}
\label{table:med-cls}
\begin{tabular}{@{}l|l|l|c@{}}
\toprule
Dataset                   & Modalities                    & Methods                           & Accuracy                      \\ \midrule
                          & Image-Only                    & ResNet-101 \cite{he2016resnet}    & 63.84                         \\ \cmidrule(l){2-4} 
                          &                               & LSTM \cite{hochreiter1997lstm}    & 59.20                         \\
                          &                               & RoBERTa \cite{liu2019roberta}     & 75.40                         \\
                          & \multirow{-3}{*}{Text-Only}   & SciBERT \cite{beltagy2019scibert} & 77.70                         \\ \cmidrule(l){2-4} 
                          &                               & NLF \cite{wu2021melinda}          & 76.60                         \\
                          &                               & SAN \cite{yang2016san}            & 72.30                         \\
                          &                               & ViL-BERT \cite{lu2019vilbert}     & \cellcolor[HTML]{EFEFEF}78.60 \\
\multirow{-8}{*}{MELINDA} & \multirow{-4}{*}{Multi-Modal} & Ours                              & \cellcolor[HTML]{C0C0C0}80.51 \\ \bottomrule
\end{tabular}
\end{table}
\noindent\textbf{Datasets}~
In our experiments, we perform the pre-training on three datasets, which are described as follows:
\begin{itemize}
    \item \textbf{ROCO} \cite{pelka2018roco}: a dataset of radiology figure-caption pairs from PubMed Central, an open-access biomedical literature database.
    It has over 81,000 radiology images (from various imaging modalities) and their corresponding captions.
    \item \textbf{MedICaT} \cite{subramanian2020medicat}: a dataset of medical figure-caption pairs also extracted from PubMed Central.
    Different from ROCO, 75\% of its figures are compound figures, including several sub-figures.
    It contains over 217,000 images with their captions and inline textual references.
    \item \textbf{MIMIC-CXR} \cite{johnson2019mimic}: the largest radiology dataset to date that consists of 473,057 chest X-ray images (in frontal or lateral views) and 206,563 reports from 63,478 patients from the Beth Israel Deaconess Medical Center.
\end{itemize}
For all the datasets, we exclude those samples with the length of their texts less than 3.
For ROCO and MedICaT, we filter non-radiology samples, and for MIMIC-CXR, we only keep images in the frontal view.
As for the dataset split, we adopt the official splits of ROCO and MIMIC-CXR.
For MedICaT, we randomly sample 1,000 image-text pairs for validation and 1,000 for test, and the remaining image-text pairs are used for training.

\begin{table}[t]
\centering
\caption{Results on the Med-ITR task (i.e., the ROCO dataset) to compare with the state-of-the-art methods, where the zero-shot (Ours (ZS)) and fine-tuned results (Ours (FT)) are shown.}
\label{table:med-irtr}
\begin{tabular}{@{}l|ccc|ccc@{}}
\toprule
                         & \multicolumn{3}{c|}{T2I}                                                                      & \multicolumn{3}{c}{I2T}                                                                       \\ \cmidrule(l){2-7} 
\multirow{-2}{*}{Methods}                & R@1                           & R@5                           & R@10                          & R@1                           & R@5                           & R@10                          \\ \midrule
ViT+BERT \cite{dou2021meter}             &  5.25                         & 15.85                         & 25.85                         &  6.85                         & 21.25                         & 31.60                         \\
ViLT \cite{kim2021vilt}                  &  9.75                         & 28.95                         & 41.40                         & 11.90                         & 31.90                         & 43.20                         \\
METER \cite{dou2021meter}                & 11.30                         & 27.25                         & 39.60                         & 14.45                         & 33.30                         & 45.10                         \\ \midrule
Ours (ZS)                                & \cellcolor[HTML]{EFEFEF}23.50 & \cellcolor[HTML]{EFEFEF}49.05 & \cellcolor[HTML]{EFEFEF}63.00 & \cellcolor[HTML]{EFEFEF}23.45 & \cellcolor[HTML]{EFEFEF}50.60 & \cellcolor[HTML]{EFEFEF}62.05 \\
Ours (FT)                                & \cellcolor[HTML]{C0C0C0}29.65 & \cellcolor[HTML]{C0C0C0}56.95 & \cellcolor[HTML]{C0C0C0}69.30 & \cellcolor[HTML]{C0C0C0}29.35 & \cellcolor[HTML]{C0C0C0}57.50 & \cellcolor[HTML]{C0C0C0}70.40 \\ \bottomrule
\end{tabular}
\end{table}

\noindent\textbf{Implementation Details}~
For the uni-modal encoders, we use the vision encoder with CLIP-ViT-B \cite{radford2021clip} ($L_{v}=12$) and the language encoder with RoBERTa-base \cite{liu2019roberta} ($L_{l}=12$).
For the multi-modal fusion module, we set the number of Transformer layers $L_{m}=6$, and the dimension of the hidden states $D=768$ with the number of heads set to 12.
For knowledge representation and injection, we set the dimension of the hidden states $D_{e}=256$.
For the pretext tasks, we adopt (knowledge-enhanced) MLM, MIM \cite{he2021mae}, and ITM, where the masking ratios of MLM and MIM are set to 15\% and 75\%, respectively.
For the optimization, the models are trained with AdamW optimizer \cite{loshchilov2018adamw} for 100,000 steps with the learning rates for the uni-modal encoders and the remaining parameters set to 1e-5 and 5e-5, respectively.
The warm-up ratio is set to 10\%, and the learning rate is linearly decayed to 0 after warm-up.
Besides, we use center-crop to resize each image to the size of 288$\times$288.

\subsection{Vision-and-Language Transfer Tasks}
To evaluate the performance, we construct a medical vision-and-language understanding benchmark including three tasks.
The details of the tasks and fine-tuning strategies are described below.

\noindent\textbf{Medical Visual Question Answering (Med-VQA)}~
This task requires the model to answer natural language questions about a medical image.
We adopt three publicly available Med-VQA datasets (i.e., VQA-RAD \cite{Lau2018VQARAD}, SLACK \cite{liu2021slake} and VQA-2019 \cite{abacha2019medvqa}), where VQA-RAD consists of 315 images and 3515 questions, SLACK contains 642 images and 14,028 questions, and VQA-2019 contains 4,200 images and 15,292 questions.
To fine-tune on this task, we regard it as a multi-label classification task and feed the concatenation of the image and text representations to a two-layer MLP to predict the corresponding answer.
During training, the models are trained with a binary cross-entropy loss with a batch size of 64.

\noindent\textbf{Medical Image-Text Classification (Med-ITC)}~
This task aims to produce the classification label given an image-text pair.
We evaluate the performance on MELINDA \cite{wu2021melinda}, a Biomedical Experiment Method Classification dataset that contains 5,371 image-text pairs.
To fine-tune on this task, we learn a two-layer MLP on top of the concatenation of the image and text representations.
We train the models with a cross-entropy loss with a batch size of 16 over a maximum of 20 epochs.

\noindent\textbf{Medical Image-Text Retrieval (Med-ITR)}~
The target of this task is to calculate a similarity score between an image and a text and then perform cross-modal retrieval.
There are two subtasks for this task, where image-to-text (I2T) retrieval requires retrieving the most relevant texts from a large pool of texts given an image, and vice versa for text-to-image (T2I) retrieval.
We conduct experiments on the ROCO dataset and measure both zero-shot and fine-tuned performance.
To fine-tune on this task, we initialize the similarity score head from the pre-trained ITM head.
The model is tuned with cross-entropy loss to maximize the scores on positive pairs with 15 random texts sampled as negative samples with a batch size of 256 over a maximum of 10 epochs.
During the evaluation, we sample 2,000 image-text pairs from the ROCO test set and report the results on the sampled 2,000 image-text pairs due to the large time complexity of the ranking process.\footnote{The time complexity of the ranking process is $O(N^{2})$, where $N$ is the sample number.}

For all tasks, we use the AdamW optimizer with the learning rate set to 5e-6 and 2.5e-5 for the model backbone and prediction heads, respectively, and the warm-up ratio set to 10\%.
For the evaluation metrics, we adopt accuracy for Med-VQA and Med-ITC, and Recall@K\footnote{Recall@K corresponds to whether the ground truth is included among top K results.} (K=1, 5, 10) for Med-ITR.

\section{Experimental Results}
\subsection{Main Results}
We compare the proposed approach with existing methods on the same datasets, with their results reported in Table \ref{table:med-vqa}, \ref{table:med-cls}, and \ref{table:med-irtr}.
There are several observations drawn from different aspects.
First, our approach achieves the best performance on all tasks, which confirms the validity of the proposed pre-training approach.
Second, for Med-VQA, our approach achieves significant improvements even compared with those pre-training methods (e.g., MTPT and MMBERT), which indicates the usefulness of incorporating knowledge into the pre-training process.
Third, for Med-ITC, the results of those strong baselines (i.e., RoBERTa, SciBERT, and ViL-BERT) are achieved by continued pre-training on the MELINDA dataset.
Our approach achieves better performance without such requirements, which indicates that an appropriate design can alleviate the need for continued pre-training on downstream datasets.
Fourth, for Med-ITR, the proposed approach achieves a substantial improvement compared with the state-of-the-art methods, where ViLT and METER are two strong baselines in the general domain.
This shows that it is necessary to design an appropriate approach (including pre-training data and methods) for the medical domain.

\subsection{Ablation Studies}
To illustrate the effectiveness of our proposed approach, we perform an ablation study on the three proposed knowledge injection designs.
The experiments are conducted on the VQA-RAD dataset, and the results are reported in Table \ref{table:ablation}.

We have the following observations.
First, for the model parameters, only the RK design brings extra 12M parameters ($\sim$3.4\% of the whole model) while other designs do not add additional parameters, which justifies introducing knowledge to Med-VLP through our approach can be done with a small price.
Second, the results of pre-training with two designs (ID 5, 6, and 7) and one design (ID 2, 3, and 4) are consistently better than those of pre-training with one design (ID 2, 3, and 4) and without any knowledge injection design (ID 1), respectively.
This demonstrates the excellent compatibility and complementarity of our design perspectives, which is critical in a multi-component approach and allows us to develop more designs under such a framework.
Third, injecting the knowledge into the multi-modal fusion module (ID 3) achieves a significant improvement.
The reason behind this might be that knowledge (i.e., entities here) serves three functions:
(i) It smooths the interaction process of the image and text representations;
(ii) It provides information at a greater granularity than words;
(iii) It removes ambiguity between diverse words by linking to the knowledge base.
Fourth, performing the aligning process can further improve the performance of the RK design (ID 5), which can be explained by the fact that the aligning processing can produce better knowledge representations for the RK process.
Fifth, our full approach (ID 8) achieves the best performance, which confirms the effectiveness of the proposed framework for medical knowledge injection.

\subsection{Qualitative Analysis}
\begin{table}[t]
\centering
\caption{Ablation study on the three knowledge injection designs (i.e., aligning through knowledge (AK), reasoning using knowledge (RK), and learning from knowledge (LK)) on the VQA-RAD dataset, with the model parameters (Para.).}
\label{table:ablation}
\begin{tabular}{@{}l|ccc|c|ccc@{}}
\toprule
ID & AK         & RK         & LK         & Para. & Open                          & Closed                        & Overall                       \\ \midrule
1  &            &            &            & 350M  & 65.36                         & 84.19                         & 76.72                         \\ \midrule
2  & \checkmark &            &            & 350M  & 67.04                         & 84.98                         & 77.88                         \\
3  &            & \checkmark &            & 362M  & 65.56                         & 86.40                         & 78.10                         \\
4  &            &            & \checkmark & 350M  & 65.92                         & 85.29                         & 77.61                         \\ \midrule
5  & \checkmark & \checkmark &            & 362M  & \cellcolor[HTML]{EFEFEF}67.60 & \cellcolor[HTML]{EFEFEF}86.08 & \cellcolor[HTML]{EFEFEF}78.76 \\
6  & \checkmark &            & \checkmark & 350M  & 66.11                         & 86.40                         & 78.32                         \\
7  &            & \checkmark & \checkmark & 362M  & 65.36                         & 86.40                         & 78.05                         \\ \midrule
8  & \checkmark & \checkmark & \checkmark & 362M  & \cellcolor[HTML]{C0C0C0}67.60 & \cellcolor[HTML]{C0C0C0}86.76 & \cellcolor[HTML]{C0C0C0}79.16 \\ \bottomrule
\end{tabular}
\end{table}
\begin{figure*}[t]
\centering
\includegraphics[width=0.95\textwidth, trim=0 0 0 0]{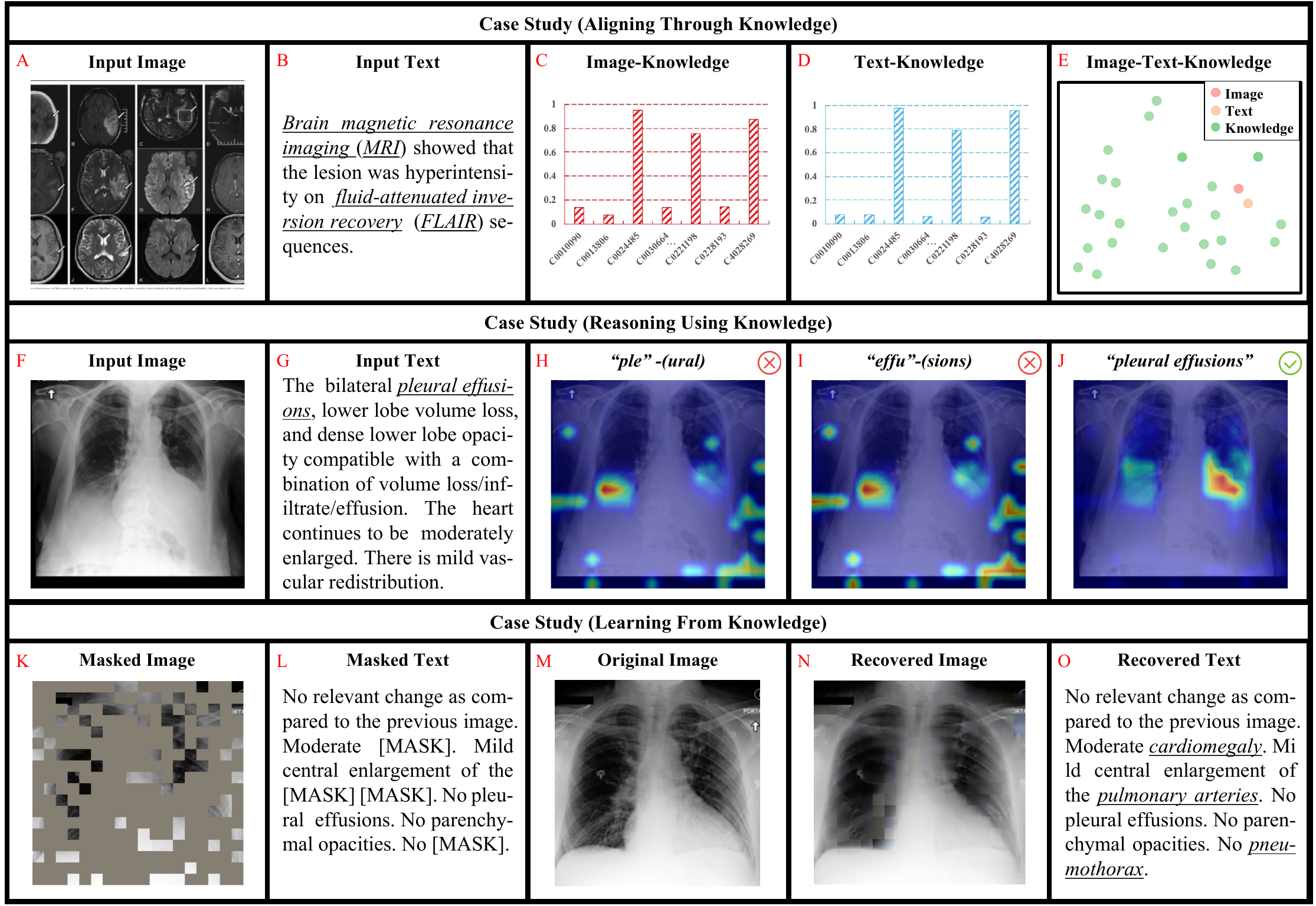}
\caption{Visualizations of the three proposed designs from the pre-trained model,
where image-knowledge and text-knowledge similarity and the image-text-knowledge t-SNE visualization are shown for "aligning through knowledge";
subword-image and entity-image attention mappings (colors from blue to red representing the weights from low to high) are shown for "reasoning using knowledge";
the recovered text (with the addition reconstructed image) is shown for ``learning from knowledge''.}
\label{fig:attention}
\end{figure*}
To further investigate the effectiveness of our approach, we perform qualitative analysis on some cases with their results shown in Figure \ref{fig:attention}.
For ``aligning through knowledge'', in the input text of this case, there is an entity ``\textit{Brain magnetic resonance
imaging}'' which links to the entity ``\textit{C4028269: Nuclear magnetic resonance imaging brain}'' in the UMLS knowledge base.
The sub-figures (3(C) and 3(D)) show that the image and text representations produced by the uni-modal encoders have high similarity with the entity representation of ``\textit{C4028269}'', which implicitly pulls the image and text representations close (as shown in the sub-figure 3(E)).
For ``reasoning using knowledge'', the sub-figures 3(H), 3(I), and 3(J) illustrate that using entities is beneficial for aligning the text with the image, where the learned entity-image attention mappings are better than the subword-image mappings.
The reason behind this is that entities are more complete semantic units.
In contrast, words (or subwords) have a smaller granularity than entities, making the correspondences between images and texts more challenging to learn.
For ``learning from knowledge'', the masked medical entities are correctly recovered by the pre-trained model (as shown in the sub-figure 3(O)) since the knowledge-induced pretext task guides the model to put more emphasis on the medical knowledge-related information.
In addition, the masked and recovered images are also shown in the sub-figures 3(K) and 3(M), respectively, which shows the high quality of the image reconstruction.
In summary, these cases reveal that injecting knowledge through the three proposed designs is essential in modeling the hidden structures among the images and texts better to promote Med-VLP.

\section{Conclusion}
In this paper, we propose to pre-train the medical vision-and-language model with medical domain knowledge, where the knowledge is injected into the Med-VLP framework from three aspects: (i) aligning the image and text representations through knowledge before their interaction; (ii) treating knowledge as the supplementation of the input image and text to assist the reasoning during the multi-modal fusion process; (iii) utilizing knowledge to induce more sophisticated pretext tasks to guide the model put more emphasis on the critical medical information.
To perform a comprehensive evaluation and facilitate further research, we construct a medical vision-and-language understanding benchmark, including three tasks (i.e., Med-VQA, Med-ITC, and Med-ITR).
Experimental results on the downstream datasets demonstrate the effectiveness of our approach, which achieves state-of-the-art performance. 
Further analyses investigate the effects of different components in our approach and show that our approach is able to better learn the correspondences between vision and language so as to produce more generic and effective vision-and-language representations.

\section*{Acknowledgement}
This work is supported in part by the Chinese Key-Area Research and Development Program of Guangdong Province (2020B0101350001), in part by the Guangdong Basic and Applied Basic Research Foundation (2020B1515020048), in part by the National Natural Science Foundation of China (61976250), in part by the Guangzhou Science and technology project (No.202102020633), and is also supported by the Guangdong Provincial Key Laboratory of Big Data Computing, The Chinese University of Hong Kong, Shenzhen.

\bibliographystyle{ACM-Reference-Format}
\bibliography{sample-sigconf}

\end{document}